\documentclass[runningheads]{llncs}
\usepackage{graphicx}
\usepackage{orcidlink}

\usepackage{tikz}
\usepackage{comment}
\usepackage{amsmath,amssymb} 
\usepackage{color}
\usepackage{graphicx}
\usepackage[accsupp]{axessibility}  

\usepackage[misc]{ifsym}
\newcommand{\ourproposedmethod}{CEP}
\usepackage{array}
\usepackage{lipsum}
\def\onedot{.}
\def\eg{\emph{e.g}\onedot}

\def\etc{\emph{etc}\onedot} 
 
\def\etal{\emph{et al}\onedot}
\makeatletter\renewcommand\paragraph{\@startsection{paragraph}{4}{\z@}
  {.5em \@plus1ex \@minus.2ex}{-.5em}{\normalfont\normalsize\bfseries}}\makeatother

\newcommand{\PreserveBackslash}[1]{\let\temp=\\#1\let\\=\temp}
\newcolumntype{C}[1]{>{\PreserveBackslash\centering}p{#1}}

\usepackage[symbol]{footmisc}

\begin{document}
\pagestyle{headings}
\mainmatter
\def\ECCVSubNumber{2649}  

\title{Context-Enhanced  Stereo Transformer} 

\titlerunning{CSTR}
%
\author{Weiyu Guo\inst{1,2}$^{\textrm{(\Letter)}}$\orcidlink{https://orcid.org/0000-0001-5449-9490} \and
Zhaoshuo Li\inst{3}$^{\textrm{(\Letter)}}$\orcidlink{https://orcid.org/0000-0001-5874-4713} \and
Yongkui Yang\inst{1}$^{\textrm{(\Letter)}}$\orcidlink{https://orcid.org/0000-0003-1159-3115} \and
Zheng Wang\inst{1}$^{\textrm{(\Letter)}}$\orcidlink{https://orcid.org/0000-0003-2855-9570} \and
Russell H. Taylor\inst{3} \and Mathias Unberath\inst{3} \and Alan Yuille\inst{3} \and Yingwei Li\inst{3}$^{\textrm{(\Letter)}}$\orcidlink{https://orcid.org/0000-0002-0711-7004}}
\authorrunning{W. Guo et al.}
%
\institute{Shenzhen Institute of Advanced Technology, Chinese Academy of Sciences, Shenzhen, China\and
University of Chinese Academy of Sciences, Beijing, China\\
 \and
Johns Hopkins University, Baltimore, USA\\
\email{\{wy.guo,yk.yang,zheng.wang\}@siat.ac.cn, \{zli122, yingwei.li\}@jhu.edu} }
\maketitle

\begin{abstract}
Stereo depth estimation is of great interest for computer vision research. However, existing methods struggles to generalize and predict reliably in hazardous regions, such as large uniform regions. To overcome these limitations, we propose Context Enhanced Path (\ourproposedmethod{}). 
\ourproposedmethod{} improves the generalization and robustness against common failure cases in existing solutions by capturing the long-range global information. We construct our stereo depth estimation model, Context Enhanced Stereo Transformer (CSTR), by plugging \ourproposedmethod{} into the state-of-the-art stereo depth estimation method Stereo Transformer. CSTR is examined on distinct public datasets, such as Scene Flow, Middlebury-2014, KITTI-2015, and MPI-Sintel. We find CSTR outperforms prior approaches by a large margin. For example, in the zero-shot synthetic-to-real setting, CSTR outperforms the best competing approaches on Middlebury-2014 dataset by 11$\%$. Our extensive experiments demonstrate that the long-range information is critical for stereo matching task and \ourproposedmethod{} successfully captures such information\footnote[2]{Code available at: \href{https://github.com/guoweiyu/Context-Enhanced-Stereo-Transformer}{github.com/guoweiyu/Context-Enhanced-Stereo-Transformer}}.
\end{abstract}

\section{Introduction}

\begin{figure*}[t]

  \centering
  
   \includegraphics[width=1\linewidth]{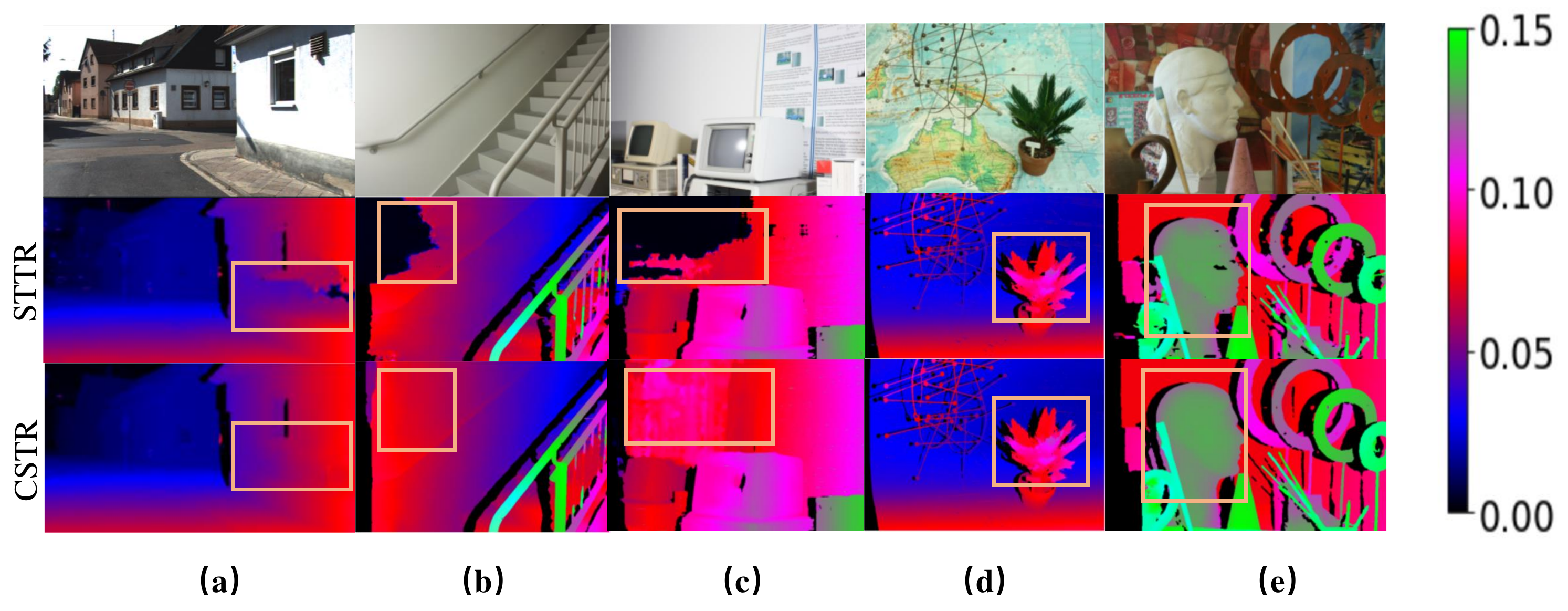}

   \caption{ Sample visualizations of hazardous regions taken from KITTI-2015 and Middlebury-2014 datasets. First row is the input left images. Second row is the  disparity predicted by Stereo Transformer (STTR) \cite{Li2020}. Third row is the disparity predicted by our proposed Context Enhanced Stereo Transformer (CSTR). The color map shown on the right is based on the disparity value relative to the image width.  }
   \label{fig:textureless}
\end{figure*}

Stereo depth estimation is a critical task in computer vision that has been widely used in various fields, such as robotics \cite{schmid2013stereo}, autonomous driving \cite{menze2015object}, and 3D scene reconstruction \cite{tomono2009robust}.
Recent developments in learning-based stereo disparity estimation algorithms generally use using techniques restricted to local information for matching the feature patterns between the left and right images. For example, prior works \cite{chang2018pyramid,guo2019group,zhang2019ga} construct a cost volume with pre-defined disparity range and use 3D convolutions to process the cost volume, limiting themselves to the receptive field of convolution kernel. Xu \etal \cite{xu2020aanet} proposed to instead process the cost volume using 2D convolutions, however, facing the same challenge. Recently, approaches that attempt to capture more global information have been proposed. For example, STTR \cite{Li2020} and RAFT-Stereo \cite{lipson2021raft} computes attention or correlation between all pixels of the left and right images on the same epipolar lines. However, they all fail to take advantage of cross-epipolar line information, which is a critical component of global information processing. Thus, as shown in Figure~\ref{fig:textureless}, these methods cannot address hazardous regions like textureless, large uniform regions, specularity, and transparency~\cite{laga2020survey,zhang2018unrealstereo}, which are particularly challenging for stereo algorithms to produce reliable estimates. The features of left and right frames in these regions are often similar or misleading, which makes the feature matching ambiguous \cite{zhang2018unrealstereo}. If disparities of these regions cannot be reliably predicted, downstream applications, such as 3D object detection \cite{sun2020disp}, may be severely impacted due to missing or wrong predictions. Therefore, in this paper, we seek to answer this critical question: how to guide the stereo models properly handle those hazardous regions.

To address this question, we hypothesize that the long-range contextual information help to improve the predictions on hazardous regions. For example, as shown in Figure~\ref{fig:textureless} (a),  previous work performs unreliably in large white wall. However, if we could use the global information (\eg, orientation, edge information) of the house, the prediction can be improved. Such global context information in theory will inform the model about the geometry on a global scale and guide the model to resolve the ambiguity in prediction.  To this end, we proposed a plug-in module, called Context Enhanced Path (\ourproposedmethod{}), which helps stereo matching models to better understand the global structure of the hazardous regions. Compared to existing methods, \ourproposedmethod{} offers the following three unique advantages:  (1) strong generalization ability, compared with previous methods~\cite{chang2018pyramid,Li2020}, \ourproposedmethod{} shows strong results on unseened real-world data even if only training on synthetic data; (2) robustness against hazardous, thanks to modeling the long-range contextual information.(3) generic, unlike~\cite{2017Learned,2018DeepMVS,yao2021decomposition}, our method serves as a plug-in that can be potentially applied to most of stereo matching methods.
We construct our stereo depth estimation model based on \ourproposedmethod{}, namely Context Enhanced Stereo Transformer (CSTR). We have examined CSTR on several popular and diverse datasets, such as, Middlebury-2014\cite{scharstein2014high}, KITTI-2015~\cite{menze2015object}, and MPI sintel~\cite{butler2012naturalistic}.
Our extensive experiments demonstrate that (1) the long-range information is critical for stereo depth estimation, (2) CSTR attains strong generalization ability, and (3) more importantly,  CSTR can better handle hazardous regions, such as texturelessness and disparity jumps (shown in Figure~\ref{fig:textureless} and Table \ref{Table4}). This result is attributed to our simple yet powerful observation: using long-range contextual information to better understand the global structure of the image can significantly help stereo depth estimation especially for those hazardous area. This result suggests that modeling long-range context information is critical for building a robust and generalizable stereo depth estimation algorithm. 

To summarize, our contributions are 3-fold: (1) we found global contextual information is critical for stereo depth estimation; (2) we design a plug-in module, Context Enhanced Path (\ourproposedmethod{}), for generic stereo depth estimation models; (3) we integrate our plug-in module and build a stereo matching model named Context Enhanced Stereo Transformer (CSTR), which achieves the state-of-the-art generalisation results on several popular datasets, including Middlebury-2014-2014\cite{scharstein2014high}, KITTI-2015~\cite{menze2015object}, and MPI-sintel~\cite{butler2012naturalistic}.

\begin{figure*}[t]
  \centering
  
   \includegraphics[width=1\linewidth]{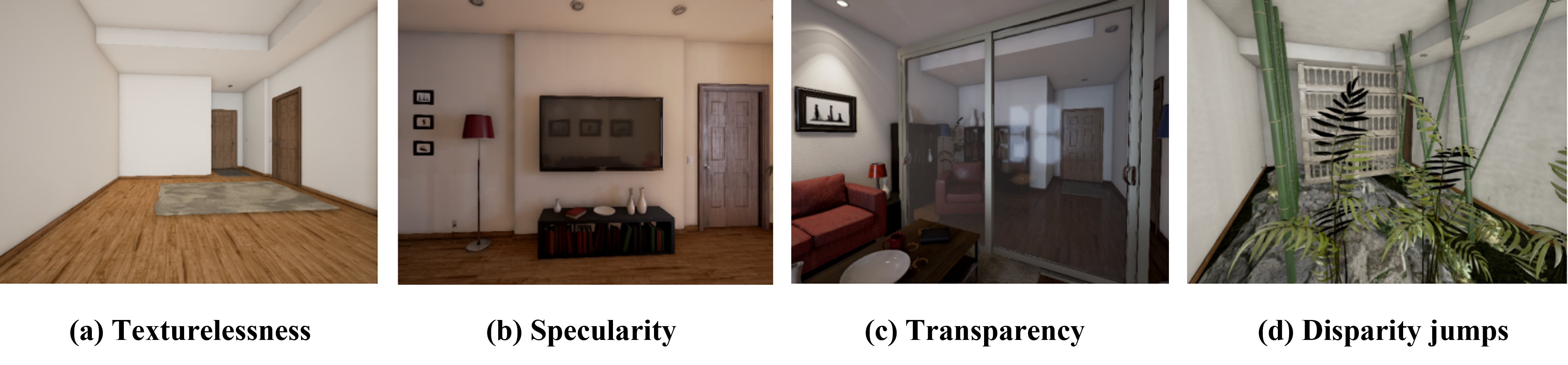}

   \caption{Examples of hazardous regions including: (a) Texturelessness: the wall and the ceiling in the room a (b) Specularity: the screen of a TV (c) Transparency: the sliding door (d) Disparity jumps: objects such as bamboos, fences and plants give frequent disparity discontinuities. Images are from Zhang~\etal~\cite{zhang2018unrealstereo}.}
   \label{fig:hazardous}
\end{figure*}
\section{Related Work}
\paragraph{Rectified stereo depth estimation} obtains per-pixel depth from the left and right frames provided by the binocular camera. It has a wide range of applications in robotics, autonomous driving, scene understanding, 3D modeling,~\etc{} In contrast to the success of deep learning in many high-level vision problems, low-level deep learning algorithms for vision tasks are still in their early stages \cite{laga2020survey}. In the field of stereo depth estimation,  many works aim to improve a single step of the classical pipeline by replacing it with a deep learning module~\cite{laga2020survey,zhao2020monocular}, where the quality of cost volume directly determines the accuracy of the disparity map. Chen~\etal~proposed Deep Embed to learn a cost function from different windows by processing multi-patches at different resolutions~\cite{chen2015deep}. After cost volume computation, cost aggregation is essential for gathering large context information from the huge cost volume. One of the most popular cost aggregation techniques is Semiglobal Matching (SGM)~\cite{hirschmuller2007stereo}. A global energy function related to the disparity map is set to minimize this energy function to solve the optimal disparity of each pixel. The raw disparity map should be refined by a post-processing algorithm. 

Although there are still several remaining challenges, recently, end-to-end deep learning begin to be used in binocular stereo depth estimation and dominate dense disparity estimation in several well-known benchmarks. In order to keep memory feasible and inference speed manageable, many researchers adopt 2D convolution-based methods.  These architectures always contain a self-design layer namely correlation layer in charge of computing correlation scores between left and right features. Mayer~$\etal$~proposed an encoder-decoder architecture based on U-net named DispNet \cite{mayer2016large}. Some researchers adopt 3D convolutions in stereo matching which take a 4D tensor (disparity range, height, weight, feature) as the input and directly process a matching volume-like representation. Chang~\etal{} proposed Pyramidal Stereo Matching network (PSMNet) to integrate Spatial Pyramidal Pooling layers (SPP) in the feature extractor \cite{chang2018pyramid}. However, these methods lead to large computational costs, such as huge memory cost and low inference speed. Besides, the disparity range of the conventional methods are limited, preventing them to be used in many cases when the scenes are close to the camera.
Recently, Li~$\etal$~use a sequence-to-sequence perspective to replace cost volume construction with dense pixel matching~\cite{Li2020}. Lipson~$\etal{}$ unify stereo and optical flow approaches and utilize GRU to iteratively generate the final disparity map~\cite{lipson2021raft}. Others~\cite{li2021temporally,li2021sins} exploit auxiliary information for detph estimation. However, stereo depth estimation is still limited by difficulties like textureless surfaces, disparity jumps, and occlusions.

\begin{figure*}[t]
  \centering
  
   \includegraphics[width=1\linewidth]{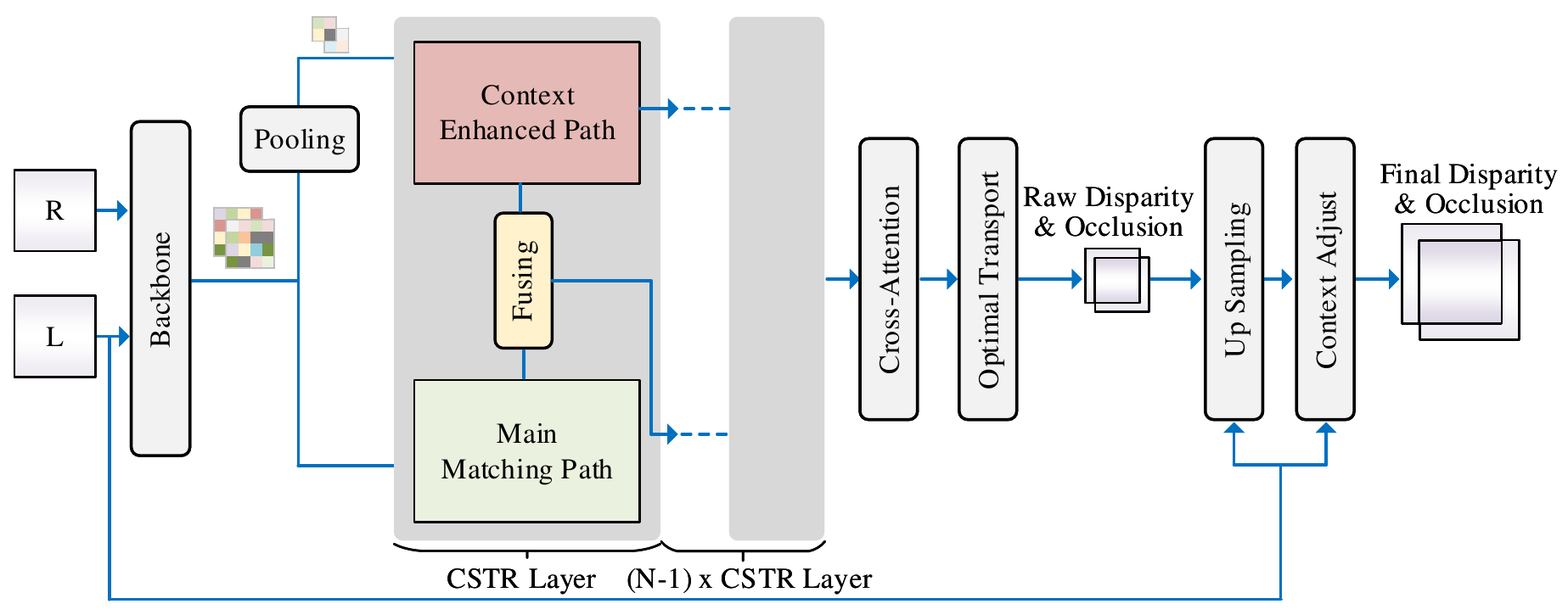}

   \caption{CSTR consists of two main components:(1) Context Enhanced Path that extracts long-range context information in low resolution feature. (2) Main Matching Path that use Axial-Attention to enhance context and Cross-Attention to compute raw disparity. Then a learnable Up Sampling block up restore the original scale of disparity and Context Adjustment block  refines the disparity with context information across epipolar lines conditioned on the left image. }
   \label{fig:network}
\end{figure*}

\paragraph{Hazardous Regions}
Most of stereo algorithms rely on the following basic assumptions~\cite{zhang2018unrealstereo}: (1) well-textured local surface  for feature extraction without large homogeneous regions; (2) single image layer assumption with only Lambertian surface; (3) the disparity varies slowly and smoothly in space without sudden jumps. However, as shown in Figure~\ref{fig:hazardous}, these assumptions can easily be broken in many real world scenarios. For example, textureless regions like large wall are commonly seen and specular surfaces will create multiple image layers. Furthermore, disparity jumps can break the local smoothness assumption. The aforementioned regions are called hazardous regions~\cite{zendel2015cv}. In this work, we specifically study these commonly seen yet challenging scenarios for more robust stereo depth estimation.

\paragraph{Efficient Attention}
Attention has a good ability to capture correspondence between two sequences and solves the problem that RNN cannot be calculated in parallel\cite{vaswani2017attention}. There are many successful applications that adopt attention to encode long-range sequences \cite{chaudhari2021attentive}. Recently, attention has been applied to extract non-local features in computer vision and led to SOTA performance for many vision tasks \cite{guo2021attention}. However, it is computational expensive when the input of attention module is large. In order to reduce its complexity, efficient attention approaches have been proposed. Yang~\etal incorporate coarse-grained global attention and fine-grained local attentions depending on the distance to the token \cite{yang2021focal} . 
\paragraph{Axial Attention}
Wang~\etal{} factorize 2D self-attention into two 1D self-attentions to propose Axial-Attention~\cite{Wang2020}. In this paper, we adopt Axial-Attention to enhance context of feature before pixel matching.
 Most previous works proposed efficient attention by adding various local constraints. 
However, these constraints always sacrifice the global context and limit the attention's receptive field.

To ensure both efficient computation and global context, Wang et al. employ two Axial-Attention layers consecutively for the height-axis and width-axis, respectively\cite{Wang2020}. A width-axis attention layer can be described as:
\begin{equation}
  y_i=\sum_{j\in N_{(W*1)}(i)}S(q_i^Tk_i+q_i^Tr_{j-i})(v_j)
  \label{eq:axial attention}
\end{equation}
 where $ N_{w*1}(i)$ is the $w*h$ scale 1D region around $i$ stands for relative position encoding, and $q,k,v,S$ denote query, key, value, soft-max, respectively. In practice, $w*h$ is much smaller than the full feature shape. 

Compared with local constraints attention, width-Axial-Attention computes the attention line by line with weight sharing. $W$ is equal to the width of input. Height-axis attention is the same as width-Axial-Attention besides computing the attention column by column.

Furthermore, positional information is critical for pixel matching, especially in large textureless regions. Due to shift-invariance in an image, we adopt relative position encoding to add data-only-dependent spatial information. A classical attention mechanism with relative position encoding can be described as follows.
\begin{equation}
\label{eq:position}
\begin{split}
a_{i,j}=x_iW_qW_k^Tx_j^T+x_iW_qW_k^Tp_j^T\\+p_iW_qW_k^Tx_j^T+p_iW_qW_k^Tp_j^T
\end{split}
\end{equation}
In Equation~\eqref{eq:position}, the four terms for addition represent content-content, content-position, position-content, position-position, respectively. However, disparity computation mainly depends on the image content. To remove redundancy and ensure efficiency, we delete the last term in Equation~\eqref{eq:position} and the equation becomes:
\begin{equation}
\label{eq:position1}
\begin{split}
a_{i,j}=x_iW_qW_k^Tx_j^T+x_iW_qW_k^Tp_j^T+p_iW_qW_k^Tx_j^T
\end{split}
\end{equation}
In the field of NLP, a similar design is adopted in DeBERT\cite{2020DeBERTa} and it is found that most tasks  only require relative position information.
\section{Context Enhanced Path}
\label{sec:CEP}
\begin{figure*}[t]
  \centering
  
   \includegraphics[width=0.9\linewidth]{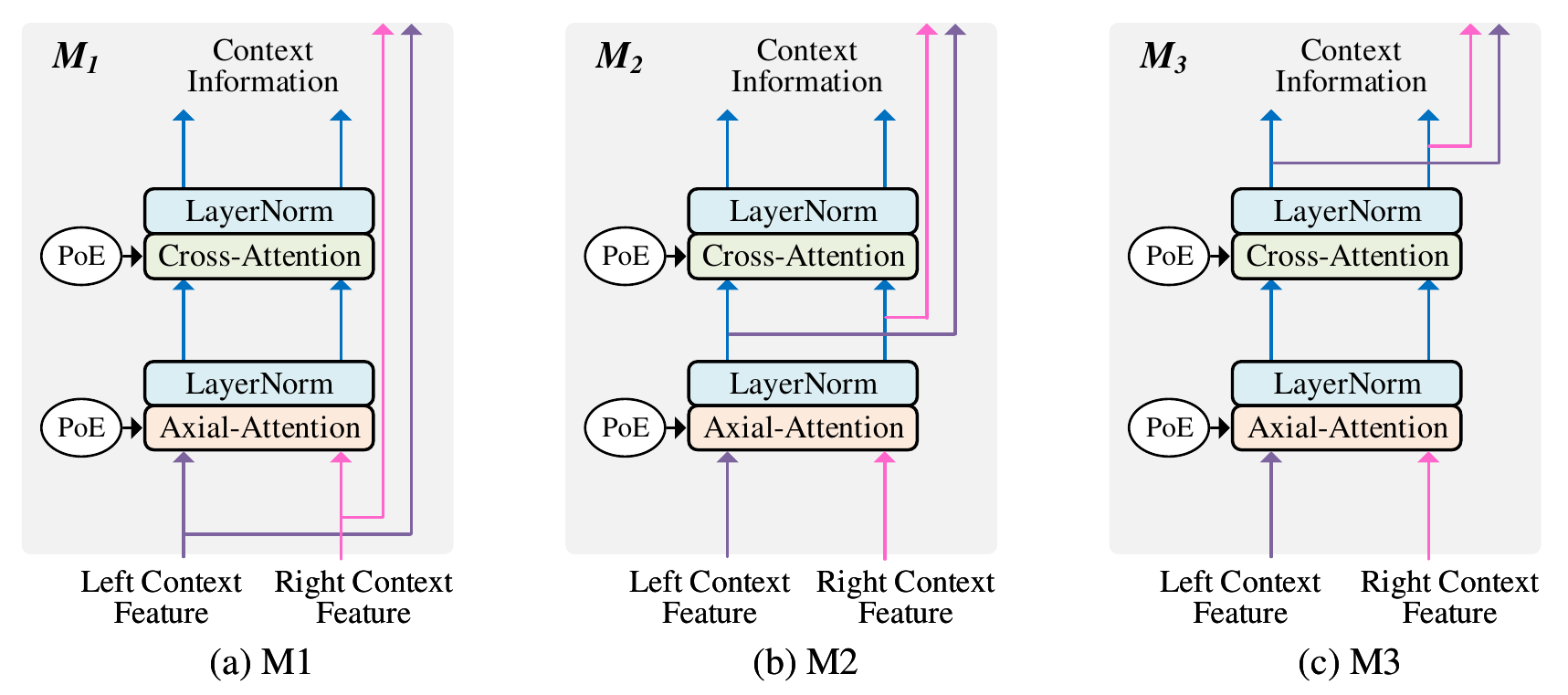}

   \caption{ Three different design choices for Context Enhanced Path (\ourproposedmethod{}). All strategies are composed of Axial-Attention and Cross-Attention, but the feature fed to next layer is different. }
   \label{fig:CEP}
\end{figure*}

We propose a plug-in module, Context Enhanced Path (CEP), that provides additional context information to help stereo matching model to better understand the global structure of the input images. The goal of \ourproposedmethod{} is to maintain the context features for left and right images, and provide the context features to the Main  Matching Path as additional complementary information. The detailed structure of the CEP is shown in Figure~\ref{fig:CEP}. As a layer-by-layer module, CEP first obtains the context feature from the previous CEP layer. Then, the Axial-Attention layer and the Cross-Attention layer are applied to further process the context features. The processed context features are served as the complementary information used for fusing with the Main Matching Path. Finally, we generate the context features as the input of the next CEP layer with 3 different strategies ($M_1$, $M_2$, $M_3$).
From $M_1$ to $M_3$, the enhancement of the context information extraction increases sequentially. In the $M_1$, we only use low-level features to extract context information. Specifically, the features output by backbone only go through one layer of Axial-Attention and one layer of Cross-Attention before fusing with the main matching path. Compared to $M_1$, $M_2$ extract higher-level context information. In the $M_2$, the features output by backbone go through $L$ layers of Axial-Attention and one layer of Cross-Attention before being fused to the $L$-th layer of the main matching path. In the $M_3$, the features output by backbone go through $L$ layers of Axial-Attention and $L$ layers of Cross-Attention before the fusion.
 

\section{Context Enhanced Stereo Transformer}
Based on our proposed Context Enhanced Path, we further propose a transformer-based stereo depth estimation model, Context Enhanced Stereo Transformer (CSTR). We will first introduce the architecture of CSTR, and then introduce each component in detail.

\subsection{Pipeline} 
The architecture of CSTR is shown in Figure~\ref{fig:network}. The whole pipeline is mainly following the architecture of STTR~\cite{Li2020} but the context information is enhanced. Given the pair of left (L) and right (R) input images, a convolution-based backbone is used to extract the left and the right features separately. The pair of left and right features then processed by several CSTR layers to obtain the disparity map with a coarse-to-fine manner. In each CSTR layer, there are 3 critical modules (Context Enhanced Path, Main Matching Path, and the path fusion module) that helps to incorporate the context information for generating better disparity map. The Context Enhanced Path is discussed in Section~\ref{sec:CEP}, the other two modules, Main Matching Path and the path fusion module, will be explained in detail in the rest part of this section. Finally, we apply several post-processing modules (\eg, optimal transport layer, upsampling layer, and context adjust layer) to obtain the final disparity.
\begin{figure*}[t]
  \centering
  
   \includegraphics[width=0.9\linewidth]{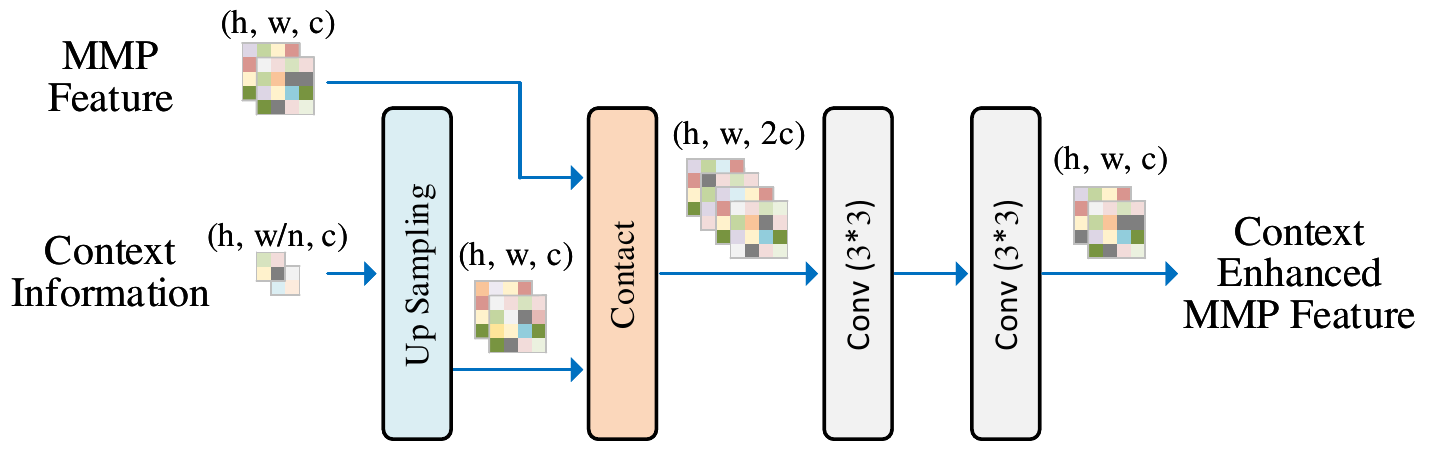}

   \caption{Path Fusion Module. (h, w,c) represent height, width and the number of feature channel. (1)The MMP feature is first concatenated with the upsampled context feature. (2) Two convolution layers are applied to the concatenated feature to aggregate the context information to the Main Matching Path(MMP) features. (3) We use the fused feature as the input of the next MMP module }
   \label{fig:fusion}
\end{figure*}
\subsection{Main Matching Path}
\begin{figure*}[t]
  \centering
  
   \includegraphics[width=0.9\linewidth]{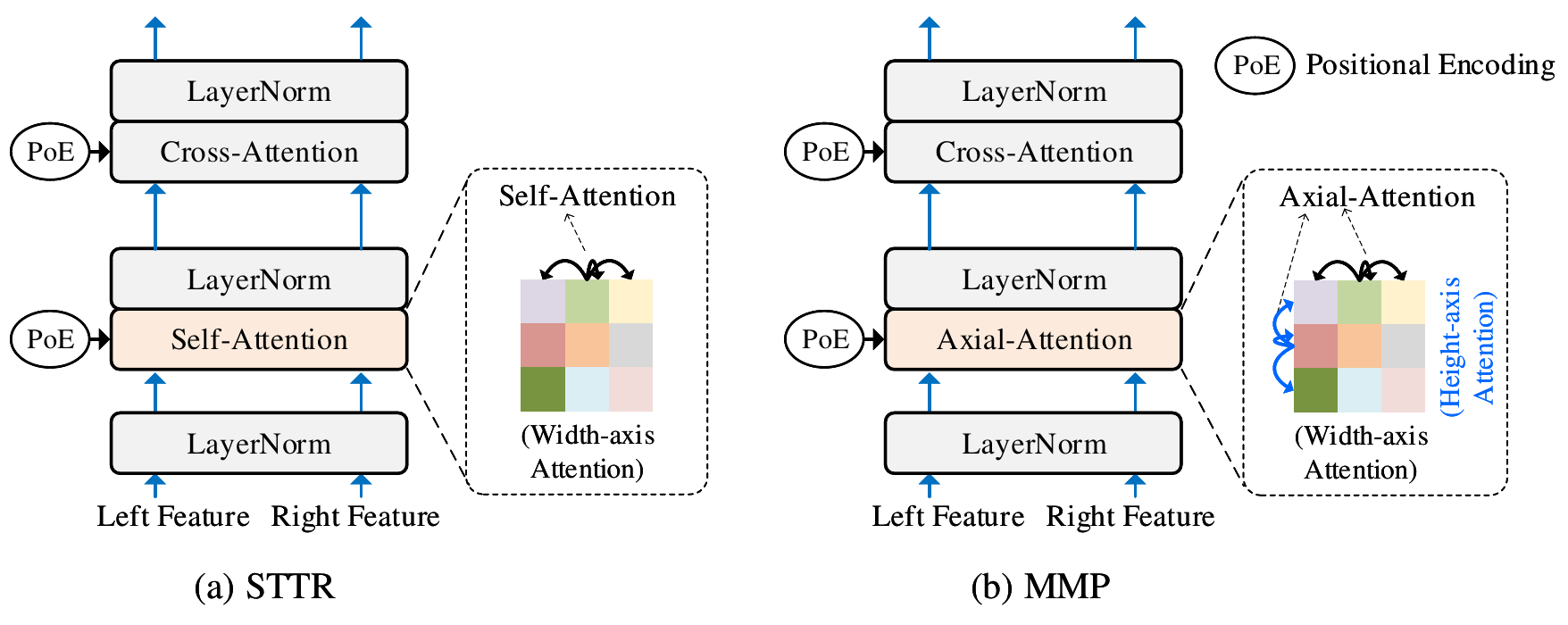}

   \caption{(a)Overview  of  the Matching Path in STTR. (b) Overview  of  the  Main Matching Path  with  alternating Axial-Attention and Cross-Attention in CSTR.}
   \label{fig:module}
\end{figure*}

Main Matching Path is similar to the Transformer module from STTR~\cite{Li2020}, which includes a self-attention module followed by a Cross-Attention module as shown in Figure~\ref{fig:module}~(a). The self-attention module is used to aggregate the information in the same image, while the Cross-Attention module is used to compute the similarity of pixels from the different images. Note that the self-attention module only computes attention between pixels along the \textit{same} epipolar line in the same image, leading to difficulty to collection contextual information from other epipolar lines.

To help the model gather more context information, as shown in Figure~\ref{fig:module}~(b), we replace the original self-attention layer to an Axial-Attention layer, including a horizontal Axial-Attention module and a vertical Axial-Attention module, to collect the context information from both horizontal and vertical axials.

\subsection{Path Fusion Module} 

The path fusion module aims to fuse the context feature from the Context Enhanced Path (CEP) to the main matching features in the Main Matching Path (MMP). This will keep the main matching path capturing long-range context from low-resolution features. The architecture is shown in Figure~\ref{fig:fusion}. Specifically, the MMP feature is first concatenated with the upsampled context feature. Then, two convolution layers are applied to the concatenated feature to aggregate the context information to the main features. Finally, we use the fused feature as the input of the next main matching path module.

\subsection{Other Important Modules}

 
This section discuss other important modules in our pipeline, including Attention Mask, Optimal Transport, Raw Disparity and Occlusion Computation. Other details are illustrated in Section~\ref{sec:Implementation Details}.
\subsubsection{Attention Mask and Optimal Transport}
We further compress the pixels' matching space based on the following two observations.

First, when a point in the physical world is imaged by a binocular camera, the imaging position in the left image will be more to the left than the imaging position in the right image. Let us denote the $P_L$, $P_R$ as the imaging point of a real point in the left and right image. Then the following formula always holds:
\begin{equation}
\label{eq:mask}
P_L-P_R<=0
\end{equation}
Therefore, the point at $P_L$ in the left image should just match the candidate point at $P>P_L$ in the right image. 

Second, every pixel in the left image can only match one pixel in the right image which is called uniqueness constraint. We adopt entropy-regularized optimal transport \cite{cuturi2013sinkhorn} to implement such constraints in a soft way. Entropy-regularized optimal transport is proposed to improve the network performance in a similar task of semantic correspondence matching \cite{liu2020semantic}. In the following section, we denote the optimal transport assignment matrix as $T$ which contains a correlation score of pixels in two images.
\subsubsection{Raw Disparity and Occlusion Computation}
In order to improve the model robustness in multi-modal distributions, we use a small number of candidate disparity in a local region rather than use all candidate disparity. First, we compute raw disparity by finding the location($S_h$) of the highest correlation score. Then,a 3 px window $N_{3x3}(S_h)$ is built around $S_h$ in matrix  $T$ to regress raw disparity. $t$ is used to represent correlation score in $N_{3x3}(S_h)$. The raw disparity regression can be described as:
\begin{equation}
\label{eq:disp0}
\sum_{i\in  N_{3*3}(S_h)}t_i = 1,i \in N_{3*3}(S_h)
\end{equation}
\begin{equation}
\label{eq:disp1}
\overline{t_i}=\frac{t_i}{\sum_{i\in  N_{3*3}(S_h)}}, i \in N_{3*3}(S_h)
\end{equation}
\begin{equation}
\label{eq:disp2}
\overline{d_{raw}}(S_h)=\sum _{i \in N_{3*3}(S_h)}d_i\overline{t_i}
\end{equation}
where $\overline{d_{raw}}$ represents regressed raw disparity and $d_i$ denotes the raw disparity in $N_{3*3}(S_h)$. Occlusion probability($ p_{occ}(S_h)$) can be interpreted as the probability that one pixel has no matching pixel in another image. Thus it can be described as:
\begin{equation}
\label{eq:occ}
p_{occ}(S_h)=1-\sum _{i \in N_{3*3}(S_h)}t_i
\end{equation}


\begin{figure*}[t]
  \centering
  
   \includegraphics[width=1\linewidth]{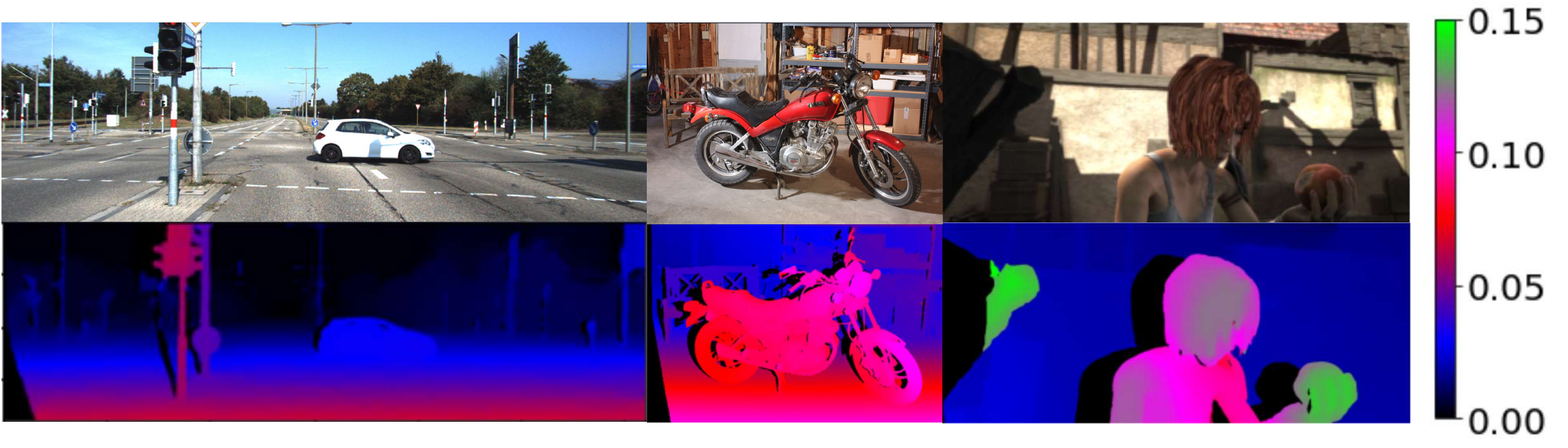}

   \caption{Results on KITTI-2015, Middlebury-2014, MPI Sintel in zero-shot synthetic-to-real setting.  Black represent occlusion.  The color map  is the image width $\times$ 0.2 and is shown on the right which used to visualize disparity.  }
   \label{fig:vis}
\end{figure*}

\section{Experiments}

\subsection{Experimental Settings}
\paragraph{Datasets.} We evaluate CSTR on four popular but diverse datasets: Scene Flow \cite{mayer2016large}, KITTI-2015~\cite{menze2015object}, Middlebury~\cite{scharstein2014high}, and MPI Sintel~\cite{butler2012naturalistic}. These datasets contain random objects, real street scene, indoor scene, and realistic artifacts, respectively.  Scene Flow is a synthetic dataset  of  random  object with many subset. We  use FlyingThings3D subset with 21818 training samples (960×540) in the experiment. KITTI-2015 contains stereo videos of road scenes from a calibrated pair of cameras mounted on a car with 200 training samples (1242×375). MPI Sintel contains sufficiently realistic scenes including natural image degradations such as fog and motion blur with 1064 training samples (1024×436).
\paragraph{Evaluation Metric.} We use both EPE (end-point-error) and 3 px Error (percentage of EPE $>$ 3)  as evaluation metrics. we use Intersection  over Union (IOU)    to evaluate occlusion estimation. In the rest of this Section, we report the results for the non-occluded regions.

\subsection{Implementation Details}
\label{sec:Implementation Details}
CSTR is implemented in Pytorch \cite{paszke2019pytorch} and is trained using one Tesla A100 GPU. During training, we use the AdamW\cite{loshchilov2017decoupled} optimizer with weight decay of 1e-4. We pre-train on Scene Flow for 17 epochs using a fixed learning rate of 1e-4 for the CSTR layer and backbone, and 2e-4 for the context adjustment layer.
\paragraph{Feature Extractor}
In order to efficiently extract both global and local context information, we adopt an hourglass-shaped feature extractor composed of encoding and decoding paths. The encoding path is based on spatial pyramid pooling~\cite{chang2018pyramid} modules and residual blocks~\cite{2016Deep} while the decoding path consists of dens-blocks~\cite{2016Densely}, transposed convolution layer, a final average pooling layer for generating multi-scale features. The scale of feature map output by transposed convolution layer is at 1/4 resolution as the input image. For an input like ($H$,$W$), we generate a multi-scale feature($H$,$W$/$2K$) with repeating average pooling of the width dimension, where $K$ is the down sample rate. 
\paragraph{Supervision}
Motivated by Relative Response loss $L_{rr}$\cite{liu2020extremely}, we split assignment matrix $T$ to matched pixel sets $M$ and unmatched pixel sets $U$. The loss can be described as:
\begin{equation}
 t_i^*=LinerInterp(T_i,p_i-d_{gt,i})
  \label{eq:loss1}
\end{equation}
\begin{equation}
L_{rr}=\frac{1}{N_M}\sum _{i \in M}-log(t_i^*)+\frac{1}{N_U}\sum _{i \in M}-log(t_{i,\Phi })
  \label{eq:loss2}
\end{equation}
where $t_i$ stands for $i$-th matching probability and $d_{gt,i}$ represents $i$-th ground truth disparity. To accelerate the convergence of the model, we adopt smooth L1 ~\cite{girshick2015fast} on both raw and final disparities. Furthermore, we use a binary-entropy loss to supervise the occlusion map. The total loss $L$ is computed as:
\begin{equation}
\begin{split}
    L=w_1L_{rr,raw}+w_2L_{d1,raw}+\\ 
    \qquad w_3L_{d1,final}+w_4L_{be,final}
\end{split}
\label{eq:loss3}
\end{equation}
where $L_{rr,raw}$, $L_{d1,raw}$, $L_{d1,final}$, $L_{be,final}$  represent Relative Response loss on raw disparity, L1 loss on  raw disparity, L1 loss on  final disparity,binary-entropy loss on final occlusion, respectively.
\begin{figure*}[t]
  \centering
  
   \includegraphics[width=1\linewidth]{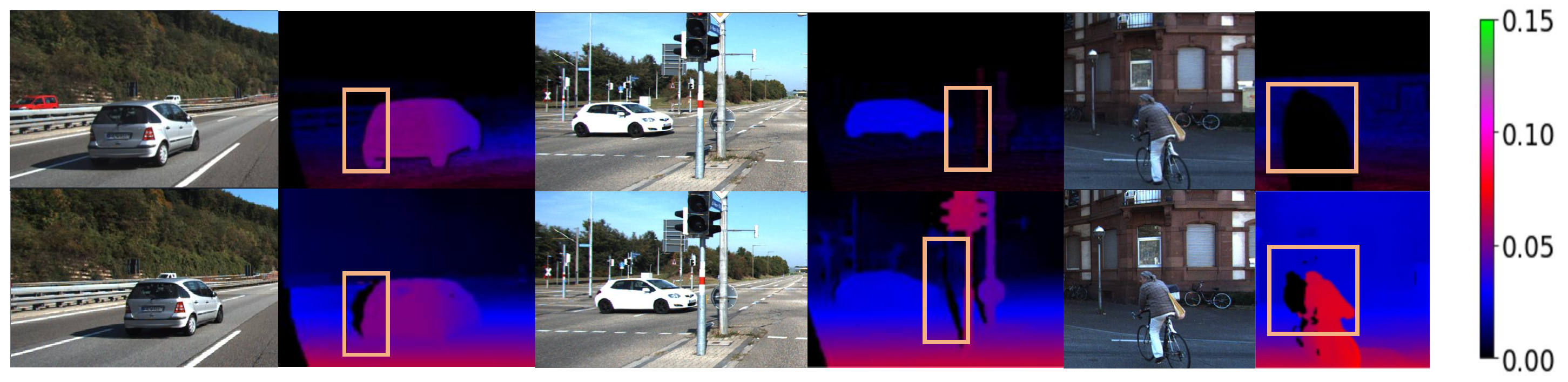}

   \caption{ KITTI-2015's ground truth is missing part of occlusion. However CSTR can accurately give this part of the missing occlusion. First row are left image and ground truth. Second row are right image and our predicted disparity.  }
   \label{fig:occ}
\end{figure*}
\paragraph{Hyperparameters.}
In our experiments, we use 6 CSTR layers with feature of 128 channels. We use multi-head attention with 4 heads. The resolution of  feature in MMP is set to $1/4$ of full resolution. Sinkhorn algorithm is run for 10 iterations \cite{cuturi2013sinkhorn}.
 
\paragraph{Baselines.}
In this work, we compare CSTR with prior work based on different learning-based stereo depth paradigms: \textbf{PSMNet}~\cite{chang2018pyramid} is a 3D convolution-based model consists two main modules --- spatial pyramid pooling and 3D CNN; \textbf{AANet}\cite{xu2020aanet} is a correlation-based model which is proposed to replace  3D convolutions to realize fast inference speed while ensure comparable accuracy; \textbf{GwcNet-g}\cite{guo2019group} is a correlation and 3D convolution hybrid approach which constructs the cost volume by group-wise correlation; \textbf{STTR}\cite{Li2020} is a transformer-based model which revisits the problem from a sequence-to-sequence correspondence perspective to replace cost volume construction; \textbf{RAFT-stereo}\cite{lipson2021raft} is a state-of-the-art recurrent model on Middlebury-2014 and Scene Flow datasets using iterative refinement to compute disparity. 

\begin{table*}[t]
    \centering
        \caption{Generalization experiment. The models are only trained on Scene Flow without fine-tuning on MPI Sintel, KITTI-2015, Middlebury-2014 dataset.  \textbf{Bold} is the best result.   }\label{table1}
       
        \resizebox{\textwidth}{15mm}{
        \begin{tabular}{l|c|c|c|c|c|c|c|c|c}
            & \multicolumn{3}{c|}{Middlebury 2014(varies)} & \multicolumn{3}{c|}{MPI Sintel$\dagger $ (1024 * 436)}&\multicolumn{3}{c}{KITTI-2015 (1242 * 375)}\\
            
            \cline{2-10}
            & 3px Error $\downarrow $ & EPE $\downarrow $ & Occ IOU $\uparrow $ & 3px Error $\downarrow $ & EPE $\downarrow $ & Occ IOU $\uparrow $ & 3px Error $\downarrow $ & EPE $\downarrow $ & Occ IOU $\uparrow $\\
            \hline
            AANet  & 6.29 & 2.24 & Null & 9.57 & \textbf{1.71} & Null & 7.06 & 1.31 & Null\\
            PSMNet & 7.93 & 3.70 & Null  & 10.24 & 2.02 & Null& 7.43 & 1.39 & Null\\
            GwcNet-g  & 5.83 & 1.32 & Null & 6.60 & 1.95 & Null& 6.75 & 1.59 & Null\\
            RAFT-Stereo & 7.57  & 1.21 & Null & 13.02 & 17.36 & Null & \textbf{5.68} & \textbf{1.10} & Null\\
            STTR & 6.19  & 2.33 & 0.95 & 5.75 & 3.01 & 0.86 & 6.74 & 1.50 & 0.98\\
            \hline
            CSTR (Ours) &\textbf{5.16}   & \textbf{1.16} & \textbf{0.95} & \textbf{5.51} & 2.58 & \textbf{0.92} & 5.78 & 1.43 & \textbf{0.98}\\
        \end{tabular}
        }
    
\end{table*}

\subsection{Zero-Shot Generalization}
We compare the zero-shot generalization ability between our proposed CSTR and previous popular stereo depth estimation methods. Specifically, the models are trained on the SceneFlow synthetic dataset, and then test on real data such as KITTI-2015 (real outdoor scene), Middlebury-2014 (real indoor scene), and MPI Sintel (Synthesized complex game scenes).

The results are shown in Table~\ref{table1}. Our model CSTR is better than our baseline method STTR~\cite{Li2020} on all datasets and on all different metrics. For example, compared with the STTR baseline, the 3px error on Middlebury dataset is improved from 6.19 to 5.16. These improvement shows that the design of context extraction of our network facilitates generalization.

Besides, our model achieves the best results on both Middlebury 2014 and MPI Sintel datasets compared with previous methods. The quantitative results on KITTI-2015 dataset is not as good as RAFT-Stereo. Compared with RAFT-Stereo, the 3px error is dropped from 5.68 to 5.78. However, by visualizing and comparing the ground-truth label and the output of CSTR, we observe that our predicted results are even more precise than the grounding truth in the occlusion areas. See Figure~\ref{fig:occ} for more details.

\begin{table}[t]
\centering
\caption{Ablation generalization experiments.The model only trained on Scene Flow without fine-tune. Following prior work,  we validate on the Scene Flow test set.  STTR: Stereo Transformer;  MMP: Main Matching Path with Aixal-Attention; $M_1$,$M_1$,$M_1$ are three different Context Enhanced Path.}\label{Table2}
\begin{tabular}{C{1cm}|C{1cm}|C{1cm}|C{1cm}|C{1cm}|C{1cm}|C{1cm}|C{1cm}|C{1cm}|C{1cm}|C{1cm}}
\hline
\multicolumn{5}{c|}{Experiment}                          & \multicolumn{3}{c|}{Scene Flow}  &
\multicolumn{3}{c}{Middlebury-2014}\\ \hline
\multicolumn{1}{c|}{STTR} &
  \multicolumn{1}{c|}{MMP} &
  \multicolumn{1}{c|}{$M_1$} &
  \multicolumn{1}{c|}{$M_2$} &
  \multicolumn{1}{c|}{$M_3$} &
  \multicolumn{1}{c|}{3px Err} &
  \multicolumn{1}{c|}{EPE} &
  \multicolumn{1}{c|}{IOU}&
  \multicolumn{1}{c|}{3px Err} &
  \multicolumn{1}{c|}{EPE} &
  \multicolumn{1}{c}{IOU} 
  
   \\ \hline
\checkmark &           &           &           &           & 1.54       & 0.50       &  0.97   & 6.93      & 2.24    & 0.95    \\
\checkmark & \checkmark &           &           &           & 1.28       & 0.43       &  0.98   & 5.55      & 2.03     & 0.95     \\
\checkmark & \checkmark & \checkmark &           &           & \textbf{1.18}       & 0.42       &   0.98 & 5.47      & 1.44     & 0.95 \\
\checkmark & \checkmark &           & \checkmark &           & 1.20       & 0.42       &   0.98    & 5.38      & 1.60     & 0.95 \\
\checkmark & \checkmark &           &          & \checkmark & 1.20  & \textbf{0.42}  & \textbf{0.98}  &  \textbf{5.13}& \textbf{1.16} & \textbf{0.95}     \\
\hline
\end{tabular}%
\end{table}

\subsection{Ablations}
In Table \ref{Table2}, we  provide quantitative results for the effects of the Axial-Attention and three different context extraction strategies. All ablated models are trained on FlyingThings of Scene Flow. Below we describe each of the experiments in more detail.

\subsubsection{Main Matching Path} Main Matching Path adopt Axial-Attention which factorizing 2D self-attention into two 1D self-attentions rather than the stand-alone self attention. This allows performing attention in a larger region to extract context information with acceptable computation cost. As shown in Table \ref{Table2}, comparing with STTR which adopt 1D attention on epipolar,  Main Matching Path have better EPE and 3px Err. Especially, 
it reduce EPE by 17$\%$ and reduce 3px Err by 14$\%$ on Scene Flow.  This improvement shows that the global information extracted by Axial-Attention is benefit to stereo matching.

\subsubsection{Three Context Enhanced Path Strategies}
We design three different context enhanced strategies($M_1$,$M_2$,$M_3$) that extract the context from low resolution features to enhanced the Main Matching Path. $M_1$,$M_2$,$M_3$ improve the result of EPE and 3px Error on Scene Flow, especially, $M_3$  achieves an EPE  7$\%$ reduction. As it has been approved, these context enhanced strategies of CEP further impove the stereo matching performance. The result on Scene FLow of three strategies are verly similar, we will further compare their real-world generalization performance and robustness in following Section.

 
  

\subsubsection{Real World Generalization Experiment}
\label{sec:real world Genera}
We evaluate the generalization performance of MMP and three CEP on Middlebury-2014. The model only trained on FlyingThings of Scene Flow. As listed in Table~\ref{Table2}, the MMP significantly outperform the baseline setting STTR which only has 1D self-attentions epipolar. The Axial-Attention for global information extraction  reduce the EPE and the 3px Err by 20$\%$ and 9$\%$  on Middlebury-2014. This shows the global information is critical for  model's generalization.  

All three context enhanced strategies outperform the MMP. It shows that the context information provided by CET facilitates generalization. $M_3$ achieves the best results on both EPE and 3px Err. For example, compared with MMP, $M_3$ reduce the 3px error by 7$\%$ and even reduce the EPE by 42$\%$. This proves that the design of our CEP is important for enhancing generalization performance.  Finally, compared with STTR, the best setting of CSTR with Axial-Attention and CEP of $M_3$ reduce the 3px error by 26$\%$ and even reduce the EPE by 48$\%$.

\subsubsection{Robustness Against Hazardous regions}
\label{sec:Robustness Against Hazardous regions}
The images regions like texturelessness, transparency, specularity are likely to cause the failure of an algorithm, namely hazardous regions\cite{zendel2015cv}. Zhang \etal~\cite{zhang2018unrealstereo} lable  the hazardous regions in KITTI-2015 and we use it to provide quantitative results for the effects of the MMP, three CEP strategies  summarized in Table~\ref{Table4}. Using Axial-Attention instead of stand alone self-attention can effectively improve average EPE of harzardous regions in with 11$\%$, especially on Textureless regions with 17$\%$. Using $M_1$ or $M_2$, which context feature just past one Cross-Attention layer,  lead to a decrease in average EPE. However, $M_3$ which are used in our final CSTR, can bring additional 8 $\%$ improvement in average EPE compared with MMP. This may be because $M_3$ uses the same number of Cross-Attention layers as MMP, which is beneficial for MMP to better integrate context information.

\begin{table}[t]
\centering
\caption{EPE results of Ablation and Rubostness experiments.The model only trained on Scene Flow without fine-tune. Hazardous Data is a dataset that label the hazardous regions in KITTI-2015\cite{zhang2018unrealstereo}; SPL: Specularity ; TEL:Texturelessness; TRS:Transparency.}\label{Table4}
\begin{tabular}{C{1cm}|C{1cm}|C{1cm}|C{1cm}|C{1cm}|C{1.5cm}|C{1.5cm}|C{1.5cm}|C{1.5cm}}
\hline
\multicolumn{5}{c|}{Experiment}                & \multicolumn{4}{c}{Hazardous Data}\\ \hline
\multicolumn{1}{c|}{STTR} &
  \multicolumn{1}{c|}{MMP} &
  \multicolumn{1}{c|}{$M_1$} &
  \multicolumn{1}{c|}{$M_2$} &
  \multicolumn{1}{c|}{$M_3$} &
  \multicolumn{1}{c|}{SPL} &
  \multicolumn{1}{c|}{TEL} &
  \multicolumn{1}{c|}{TRS} &
  \multicolumn{1}{c}{AVG} 
   \\ \hline
\checkmark &           &           &           &               & 5.43     & 10.42    & 7.03 & 7.63   \\
\checkmark & \checkmark &           &           &                & 4.98     &8.59    & 6.8  & 6.79   \\
\checkmark & \checkmark & \checkmark &           &              & 4.75      & 10.74    & 7.64   & 7.71   \\
\checkmark & \checkmark &           & \checkmark &             & \textbf{3.68}      & 11.28   &  7.01 & 7.32\\
\checkmark & \checkmark &           &          & \checkmark & 4.54 & \textbf{8.21} & \textbf{6.01}   & \textbf{6.25}
\\
\hline
\end{tabular}%
\vskip -3ex
\end{table}

\section{Conclusions}
Current stereo depth estimation models usually fail to handle the hazardous regions. In this paper, we found using global context information mitigate this issue. Therefore, we proposed a plug-in module, Context Enhanced Path. Based on \ourproposedmethod{}, we then built a stereo depth estimation model, Context Enhanced Stereo Transformer. According to our experimental results, our method achieves strong cross dataset generalization ability, handles hazardous regions robustly, and provides accurate occlusion prediction.

\vskip 1ex
\footnotesize
\noindent\textbf{Acknowledgments} This paper is supported by Key-Area Research and Development Program of Guangdong Province (Grant No. 2019B010155003), Guangdong Basic and Applied Basic Research Foundation (Grant No. 2020B1515120044, 2020A1515110495), Johns Hopkins University internal funds, ONR award N00014-21-1-2812, and NIH award K08DC019708.

\clearpage
%
%
\bibliographystyle{splncs04}
\bibliography{egbib}
\end{document}